\documentclass[letterpaper]{article} 
\usepackage{aaai2026}  
\usepackage{times}  
\usepackage{helvet}  
\usepackage{courier}  
\usepackage[hyphens]{url}  
\usepackage{graphicx} 
\usepackage{natbib}  
\usepackage{caption} 
\usepackage{algorithm}
\usepackage{algorithmic}

\usepackage{bm}
\usepackage{tabularx}
\usepackage{booktabs}
\usepackage{multirow}
\usepackage{amsmath}
\usepackage{multicol}
\usepackage{xcolor}
\usepackage{caption}
\usepackage{amssymb}
\usepackage{lipsum}
\usepackage{cuted}
\usepackage{newfloat}
\usepackage{listings}
\DeclareCaptionStyle{ruled}{labelfont=normalfont,labelsep=colon,strut=off} 
\floatstyle{ruled}
\newfloat{listing}{tb}{lst}{}
\floatname{listing}{Listing}
\title{SparseWorld: A Flexible, Adaptive, and Efficient 4D Occupancy World Model Powered by Sparse and Dynamic Queries}
\author{
    Chenxu Dang\textsuperscript{\rm 1,2}\thanks{ Working done during the internship at AIR.},
    Haiyan Liu\textsuperscript{\rm 3},
    Jason Bao\textsuperscript{\rm 3},
    Pei An\textsuperscript{\rm 1},
    Xinyue Tang\textsuperscript{\rm 3},
    PanAn\textsuperscript{\rm 4},
    Jie Ma\textsuperscript{\rm 1}\thanks{Corresponding Authors.},\\
    Bingchuan Sun\textsuperscript{\rm 3}\footnotemark[\value{footnote}],
    Yan Wang\textsuperscript{\rm 2}\footnotemark[\value{footnote}]\\
}

\affiliations{
    \textsuperscript{\rm 1} Huazhong University of Science and Technology\\
    \textsuperscript{\rm 2} Institute for AI Industry Research (AIR), Tsinghua University\\  
    \textsuperscript{\rm 3} Lenovo Group Limited\\
    \textsuperscript{\rm 4} AIR Wuxi Innovation Center, Tsinghua University (AIRIC)

}

\usepackage{bibentry}

\begin{document}

\maketitle

\begin{abstract}
Semantic occupancy has emerged as a powerful representation in world models for its ability to capture rich spatial semantics. However, most existing occupancy world models rely on static and fixed embeddings or grids, which inherently limit the flexibility of perception. Moreover, their ``in-place classification" over grids exhibits a potential misalignment with the dynamic and continuous nature of real scenarios.
In this paper, we propose \textbf{SparseWorld}, a novel 4D occupancy world model that is flexible, adaptive, and efficient, powered by \textbf{sparse} and \textbf{dynamic} queries. We propose a Range-Adaptive Perception module, in which learnable queries are modulated by the ego vehicle states and enriched with temporal-spatial associations to enable extended-range perception. To effectively capture the dynamics of the scene, we design a State-Conditioned Forecasting module, which replaces classification-based forecasting with regression-guided formulation, precisely aligning the dynamic queries with the continuity of the 4D environment. In addition, We specifically devise a Temporal-Aware Self-Scheduling training strategy to enable smooth and efficient training.
Extensive experiments demonstrate that SparseWorld achieves state-of-the-art performance across perception, forecasting, and planning tasks. Comprehensive visualizations and ablation studies further validate the advantages of SparseWorld in terms of flexibility, adaptability, and efficiency.
\end{abstract}
\begin{links}
\link{Code}{https://github.com/MSunDYY/SparseWorld}
\end{links}

\section{Introduction}

In recent years, vision-centric end-to-end autonomous driving, which predicts the ego vehicle’s future trajectory from monocular or multi-view images, has gained significant attention. Among them, occupancy-based world models~\cite{occworld,occllama,preworld} leverage semantic occupancy representations for rich spatial understanding and have shown superior planning performance.

Early occupancy world models~\cite{occworld,dome}, as shown in Fig.~1(a), independently encode each frame’s occupancy into embeddings, which are later fused and decoded by a future-oriented world model. Such decoupled designs separate forecasting from perception, hindering gradient flow and end-to-end optimization, while the repeated encode–decode process inevitably loses fine-grained information, no matter how carefully it is designed.
\begin{figure}
    \centering
    \includegraphics[width=1.0\linewidth]{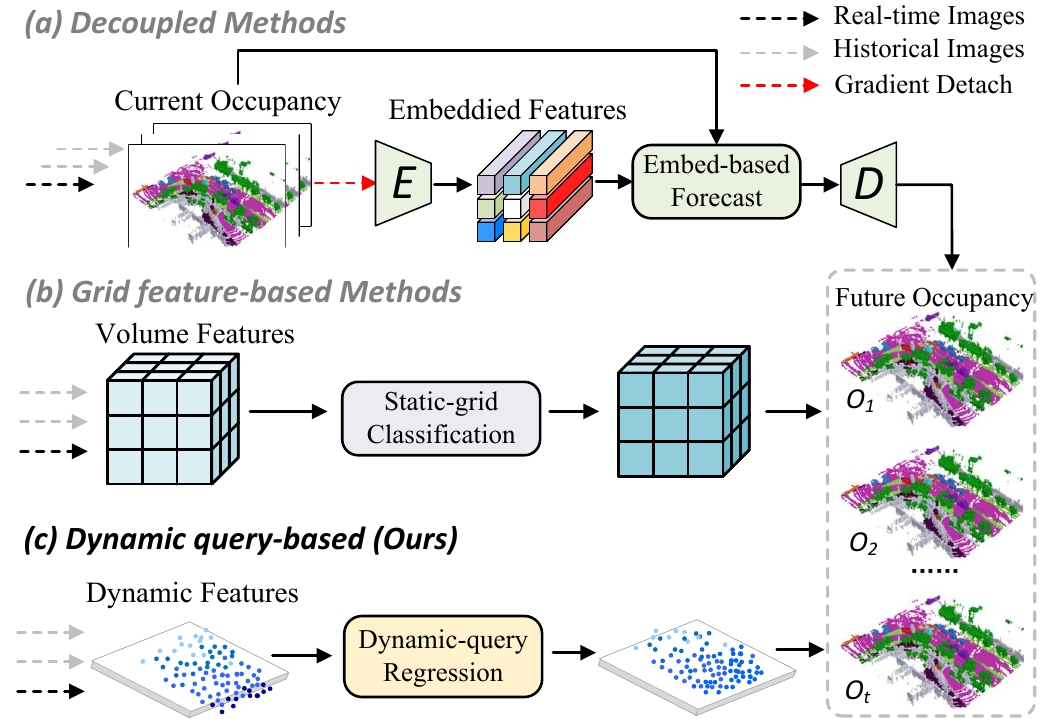}
    \caption{\textbf{(a)} Perception-forecasting-decoupled methods. \textbf{(b)} Grid feature-based methods. \textbf{(c)} We adopt dynamic query representations, which facilitate continuous and coherent 4D scene forecasting and planning.}
    \label{fig:enter-label}
\end{figure}

Recently, a series of grid feature-based works \cite{preworld,driveoccworld} have adopted perception features as intermediate representations for per-grid forecasting, as show in Figure 1(b), thus enabling end-to-end optimization. However, their static ``in-place classification" operation misaligns with the continuity of ego-motion and scene dynamics, causing temporal inconsistency, feature drift, and cumulative errors.

Both aforementioned world models rely on manually predefined spatial ranges, limiting perceptional flexibility and adaptivity. In real-world driving scenarios, where vehicle speed varies drastically, adaptive perception ranges are critical for accurate forecasting and planning. Moreover, dense grids incur high computational and memory costs while ignoring the inherently sparse nature of the physical world.

To overcome these limitations, inspired by recent sparse occupancy perception models \cite{opus,sparseocc}, we propose SparseWorld, a fully sparse 4D occupancy world model built on sparse and dynamic queries, as show in Figure 1(c). Following a ``perceive-then-forecast" paradigm, SparseWorld adaptively constructs extended-range occupancy queries, and regresses the future motion of scene elements relative to the ego vehicle, moving beyond static grid classification.

We introduce a \textbf{Range-Adaptive Perception} module that takes learnable queries as input and employs stacked decoders featuring temporal-spatial fusion. To fully enhance the perception adaptability, an Adaptive Scaling sub-module encodes ego vehicle's historical trajectory to modulate the initial distribution of queries. Leveraging the continuity and dynamics of queries, we further design a \textbf{State-Conditioned Forecasting} module, where the ego query interacts with scene queries via spatial modulation,and regression-guided migration replaces traditional ``in-place classification" to for continuous and plausible motion forecasting.

We further introduce a targeted \textbf{Temporal-Aware Self-Scheduling} training strategy that implicitly partitions query timestamps, allowing the model to autonomously learn timestamp assignments during training, which significantly improves training efficiency and autonomy.

To validate the effectiveness of SparseWorld, we conducted extensive experiments on the Occ3d-nuScenes \cite{occ3d} benchmark, comparing it with other state-of-the-art methods. The experimental results demonstrat that SparseWorld significantly outperforms dense models in both forecasting and planning tasks. Specifically, SparseWorld surpasses PreWorld \cite{preworld} by 20\%–40\% mIoU in future occupancy forecasting, and reduces collision rate in trajectory planning by half. Moreover, our SparseWorld achieves an approximate 7x speedup in inference compared to dense methods, greatly enhancing its practicality for real-world deployment.

Our main contributions can be summarized as follows:
\begin{itemize}
    \item We propose a sparse 4D occupancy world model powered by sparse and dynamic queries for flexible, adaptive, and efficient modeling of autonomous driving scenarios.
    \item We propose a Range-Adaptive Perception module featuring the ego vehicle' state and introduce a regression-oriented State-Conditioned Forecasting paradigm that effectively exploits the spatiotemporal continuity to improve 4D scene evolution forecasting.
    \item We develop a novel Temporal-Aware Self-Scheduling training strategy to enable smooth and efficient training.
    \item Our SparseWorld significantly outperforms state-of-the-art methods in both effectiveness and efficiency. We design comprehensive ablation studies, complemented by visualizations, to support our claims.
\end{itemize}
\section{Related Work}
\subsection{3D Occupancy Prediction}

Occupancy perception methods can be broadly categorized into dense and sparse paradigms according to their modeling strategies. Dense ones \cite{cotr,flashocc,bevdet4d,heightocc,protoocc,panoocc,lightocc,protoocc,cvtocc,fbocc} typically construct BEV or volume features that are conceptually straightforward but incur significant computations and limited flexibility. In contrast, sparse approaches \cite{sparseocc,opus,osp,voxformer} eliminate the reliance on dense representations. Recently, some weakly- and self-supervised approaches \cite{renderocc,selfocc,gaussianflowocc,sun2024gsrender} has emerged to alleviate the reliance on expensive 3D annotations.

\subsection{4D Occupancy World Models}
World models aim to predict future scenes and plan ego-agent trajectories based on historical observations and actions \cite{worldmodels,magicdrive,vidar,vista}. Occupancy world models are required to simultaneously forecast future occupancy scenes and plan trajectories. Early works such as OccWorld \cite{occworld} and its variants \cite{occllama,occllm} decouple occupancy perception and prediction by first encoding the observed scene and then forecasting autoregressively followed by re-encoding and decoding. Recent grid-based methods attempt to unify perception and prediction by constructing volumetric \cite{preworld} or BEV \cite{driveoccworld} features to represent the spatiotemporal world consistently. 


\subsection{End to End Autonomous Driving}
Planning-oriented end-to-end autonomous driving typically requires precise environmental perception. SP-T3 \cite{stp3} and UniAD \cite{uniad} represent the scene with a unified BEV representation, while VAD \cite{vad} introduces a vectorized design. \citet{sparsead} and \citet{sparsedrive} perform perception and planning sequentially based on sparse perception signals, whereas the more recent DriveTransformer \cite{drivetransformer} decouple and execute perception and planning in parallel, demonstrating superior performance in closed-loop inference. Some recent works have explored the use of anchor-based\cite{vadv2} and diffusion\cite{diffusiondrive,diffusion-planner} to encourage diverse trajectory outputs. In our work, we adopt autoregressive generation with L1 supervision for trajectory prediction for fair comparison.

\begin{figure*}
    
    \includegraphics[width=1.0\linewidth]{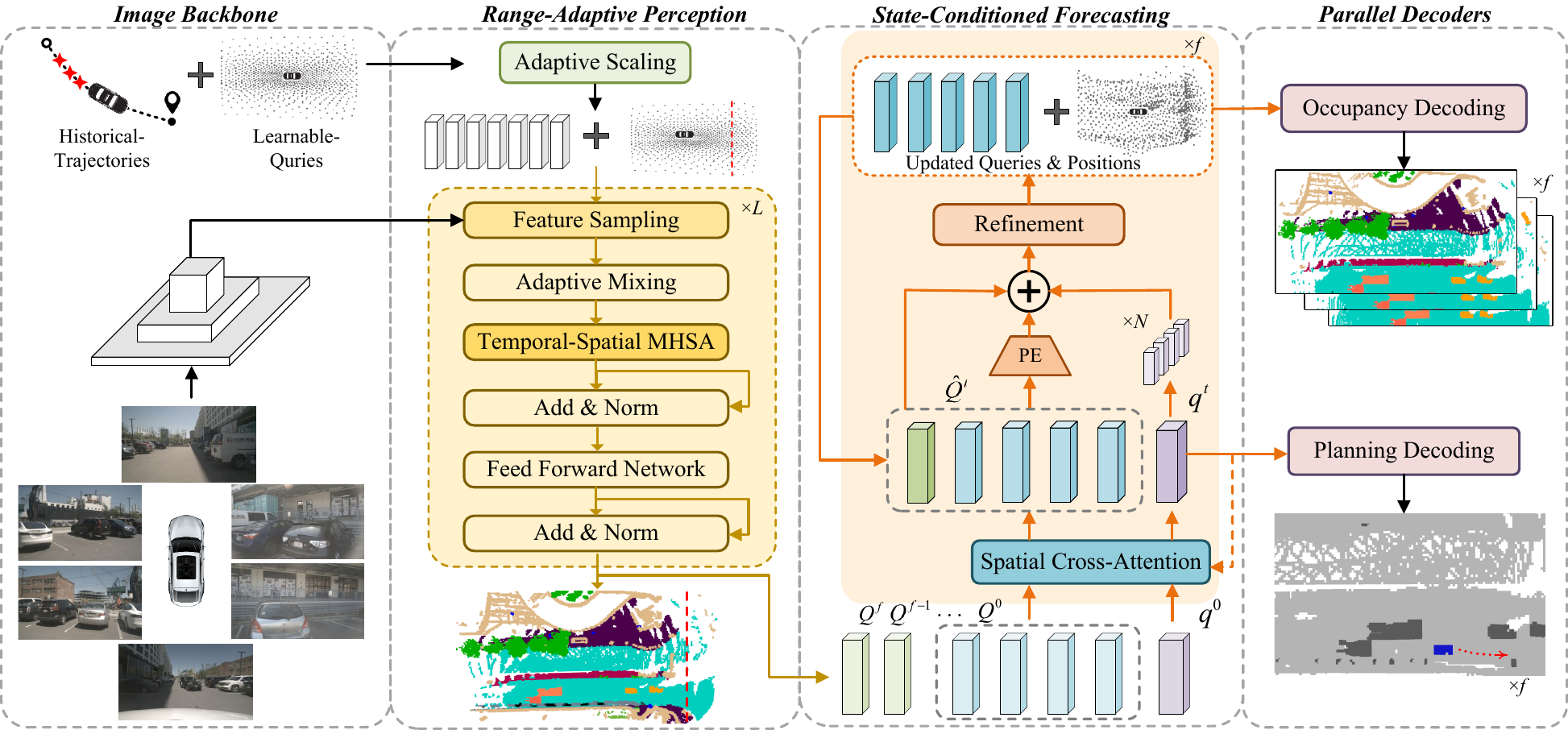}
    \caption{The overall architecture of \textbf{SparseWorld}. A set of learnable queries encoded with the historical trajectory through an Adaptive Scaling are then fed into stacked spatio-temporal decoders, which interacts with multi-frame, multi-view image features via a Temporal-Spatial MHSA. Subsequently, the extended-range queries are dynamically refined through a State-Conditional Forecasting module, which refines their positions while guiding the ego-vehicle’s motion state in an autoregressive manner. Finally, two parallel decoders are employed for forecasting and planning, respectively.}
    \label{fig:overview}
\end{figure*}

\section{Methodology}
\subsection{Preliminary}
\label{method:preliminary}
A practical and coherent AD world model is expected to take advantage of current and past $p$ frames of the ego vehicle waypoints $\{w^{i}\}_{i=-p}^0$ and sensor inputs $\{s^{i}\}_{i=-p}^0$ to predict the current and future $f$ frames of the semantic occupancy representations $\{o^i\}_{i=0}^{f}$ and the corresponding planning waypoints $\{w^{i}\}_{i=0}^{f}$. Decoupled methods embed historical occupancy observations $\{o^i\}_{i=-p}^{0}$ into compact latent embeddings and forecast the corresponding future latent codes, followed with occupancy decoders to reconstruct future occupancy states:
\begin{equation}
    z^{t} = \mathcal{E}(\textit{o}^{t}), z^{t+1}=\mathcal{W}(z^{t},z^{t-1}...), \textit{o}^{t+1} = \mathcal{D}(z^{t+1}).
\end{equation}
Here, $\mathcal{E}, \mathcal{D}, \mathcal{W}$ denote occupancy encoder, occupancy decoder, and latent embedding forecaster, respectively. In contrast, grid-based methods employ static volume features to perform per-voxel ``in-place classification'' forecasting:
\begin{equation}
    F^{0} = \mathcal{N}(s^{-p},...,s^{0}), F^{t}=\mathcal{F}(F^{t-1}), \textit{o}^{t}=\mathcal{H}(F^{t})
\end{equation}
Here, $F$ denotes the volume feature, while $\mathcal{N}$, $\mathcal{F}$, and $\mathcal{H}$ represent the perception, forecasting and occupancy head modules, respectively.

As illustrated in Figure \ref{fig:overview}, the overall architecture of SparseWorld consists of four components: (1) a generic image backbone that extracts multi-scale visual features over multiple frames; (2) a Range-Adaptive Perception module (\textbf{RAP}) that is composed of $L$ decoder layers and detailed in Section 3.2; (3) a State-Conditioned Forecasting module (SCF) that is described in Section 3.3; (4) parallel decoding heads for 4D occupancy forecasting and motion planning. In Section 3.4, we will detail our Temporal-Aware Self-Scheduling training strategy.

\subsection{Range-Adaptive Perception}
\label{method:perception}
The core of world models lies in forecasting the dynamics of the surrounding scene under the ego agent’s motion conditions. However, existing grid-based perception models, constrained by fixed spatial resolution and truncated receptive fields, are inherently limited in handling dynamic scenarios.

To address these limitations, we adopt a more flexible dynamic query formulation. As shown in Figure \ref{fig:overview}, the input of RAP consists of learnable query embeddings and corresponding 4D coordinates ($x, y, z$, timestamp). Compared to dense grids, sparse queries offer multiple advantages: (1) \textbf{Flexible}: The adaptive spatial distribution enables ultra-range perception. (2) \textbf{Continuous}: It aligns with the spatiotemporal dynamics of real-world scenarios. (3) \textbf{Efficient}: It significantly reduces the storage and computational costs. 

Following the driving intuition that faster speeds require longer perception ranges, the perception range should adapt to the ego vehicle's speed. We design an ego-guided Adaptive Scaling module to flexibly scale the perception range. Specifically, we encode historical ego waypoints $\{w^{i}\}_{i=-p}^0$, which implicitly reflect the current velocity, to modulate the initial coordinates $P_0=\{\mathbf{p}_i\}_{i=1}^{N}$ of queries $Q_0$:
\begin{equation}
    \begin{aligned}
&\boldsymbol{\gamma}=[\gamma_x,\gamma_y,\gamma_z] = \text{MLP}([w^{-p},...,w^{-1},w^{0}])\\
&\mathbf{p}'_i = \boldsymbol{\gamma} \odot \mathbf{p_i}=[\gamma_xx_i,\gamma_yy_i, \gamma_zz_i]
\end{aligned}
\end{equation}
Here, $N$ denotes the number of queries and $P_0$ is learnable, which will be explained later. The initial queries $Q_0$ combined with modulated positions $P_0'={\{\mathbf{p}_i'\}}_{i=1}^N$, regardless of timestamps, are fed info $L$ stacked decoders for extended-range occupancy perception for current moment.



\paragraph{The details of Decoder}
Our stacked decoders follow a coarse-to-fine paradigm. Each query samples semantic information from multi-view, multi-scale feature maps extracted by the image backbone, followed by adaptive mixing introduced in \cite{sparsebev,sparseocc}. The queries are then fed into Temporal-Spatial Multi-Head Self-Attention (TS-MHSA) for spatiotemporal interaction.

 The attention weights of Temporal-Spatial MHSA are composed of three components: \textbf{semantic similarity}, \textbf{spatial proximity}, and \textbf{temporal causality}.
Formally, for any two queries $q_i,q_j$ with 4D coordinates $(x_i,y_i,z_i,t_i)$ and $(x_j,y_j,z_j,t_j)$, the attention score $A_{ij}$ is computed as:
\begin{equation}
    A_{ij} = q_i^\top q_j-\tau_i||\bm{p_i}-\bm{p_j}||^2+M_{ij}
\label{eq: attention weight}
\end{equation}
where $\bm{p_i}=(x_i,y_i,z_i)$, $\tau_i$ is a learnable scaling factor controlling spatial bias inspired by SparseBEV \cite{sparsebev}, and $M_{ij}$ is a temporal masking term defined as:
\begin{equation}
M_{ij} =
\left\{
\begin{array}{ll}
0, & \text{if } t_i \ge  t_j \\
-\infty, & \text{otherwise}
\end{array}
\right.
\end{equation}
This design ensures the temporal consistency by avoiding temporal interference.

Each decoder layer is followed by an occupancy head that outputs multiple points per query. Across decoder layers, we gradually increase the number of output points and update the query positions based on the mean of the output points. From the final decoder layer, we obtain the extended-range scene queries $Q=\{q_i\}_{i=1}^N$ and their updated 3D positions $P=\{p_i\}_{i=1}^{N}$ at the current timestamp. We supervise the outputs of each decoder layer individually. 

\subsection{State-Conditioned Continuous Forecasting}
\label{method:forecasting}
Leveraging the continuous and dynamic nature of queries, we design a regression-guided continuous forecasting strategy, conditioned on the ego state. The extended-range perception queries $Q$, are temporally partitioned into $Q^0,Q^1,...,Q^f$ according to their timestamps. 

Following existing methods, we encode ego state as query. At any time step $t$, the ego query $q^{t}$ interacts with the scene query $Q^t$ through spatial cross-attention for next state $q^{t+1}$. The attention weights between $q^t$ and any scene query $q_i^t$ consist of two components:
\begin{equation}
        A^t_i = (q^t)^\top q^t_i-\tau^t_i||p_i^t||^2, \quad \tau^t_i = \text{MLP}(q^t_i)
\end{equation}

Then, the dynamic scene of next frame is forecasted as:
\begin{align}
    \hat{Q}^{t+1} = [\hat{Q}^t,Q^{t+1}]+\text{PE}([P^t,P^{t+1}])+\text{repeat}(q^{t+1})
\end{align}

Here, $[\cdot , \cdot]$ denotes concatenation for next scene augmentation, PE represents the 4D position encoding, repeat denotes broadcasting the ego query to match $\hat{Q}^{t+1}$. This design allows the model to capture both the global motion of the ego vehicle and the local state of each query.

We recursively repeat the above process. At each time step $t$, $\hat{Q}^t$ is decoded for the dynamic offset for the next frame while undergoing dynamic spatial refinement. $q^t$ is synchronously decoded for planning.

In contrast to grid-based methods that formulate occupancy forecasting as recursive per-voxel classification, we reformulate the problem as a regression task to better capture the continuous evolution of both the ego vehicle and the surroundings. This paradigm shift enables smoother and more coherent spatiotemporal modeling. We will empirically demonstrate that SparseWorld effectively mitigates feature misalignment and temporal confusion commonly observed in grid-based methods.

\subsection{Timestamp-Aware Self-Scheduling Training}
\label{train}
To equip the model with extended-range perception capability, we supervise the training of RAP using a mixture of ground truths from multiple frames. Specifically, we sparsify the occupancy ground truth $\{\hat{\mathcal{G}}^t\}_{t=0}^f$ of the next $f$ frames to extract the 3D occupied voxel coordinates as point clouds. These point clouds are then united into the current timestamp through coordinate transformation. The merged point clouds are re-voxelized to obtain the ground-truth $\hat{\mathcal{G}}$. 

The 3D coordinates $P_0'$ of queries can be learned via Chamfer distance to $\hat{\mathcal{G}}$:
\begin{equation}
    \text{CD}(P_0',\mathcal{G}) = \sum_{p \in P_0'} \min_{g \in \hat{\mathcal{G}}} \|p - g\|_2^2 + \sum_{g \in \hat{\mathcal{G}}} \min_{p \in P_0'} \|p - g\|_2^2,
\label{eq: chamfer}
\end{equation}
\setlength{\tabcolsep}{5pt}
\begin{table*}[ht]

  \small
  \renewcommand{\arraystretch}{1.3}
  \makebox[\textwidth][c]
  {%
    \begin{tabular*}{\textwidth}{>{\centering\arraybackslash}m{3.0cm}|
    >{\centering\arraybackslash}m{2.0cm}|
    >{\centering\arraybackslash}m{1.0cm}
    >{\centering\arraybackslash}m{1.0cm}
    >{\centering\arraybackslash}m{1.0cm}
    >{\centering\arraybackslash}m{1.0cm}|
    >{\centering\arraybackslash}m{1.0cm}
    >{\centering\arraybackslash}m{1.0cm}
    >{\centering\arraybackslash}m{1.0cm}
    >{\centering\arraybackslash}m{1.0cm}|
    >{\centering\arraybackslash}m{1.1cm}}
      \hline
      \multirow{2}{*}{Method} & \multirow{2}{*}{Aux. Sup.} & \multicolumn{4}{c|}{mIoU $\uparrow$} & \multicolumn{4}{c|}{IoU $\uparrow$} & \multirow{2}{*}{FPS$\uparrow$} \\
      & & 1s & 2s & 3s & Avg. & 1s & 2s & 3s & Avg. & \\
      \hline
      OccWorld-T & Semantic LiDAR &4.68 &3.36 &2.63 &3.56 &9.32 &8.23 &7.47 &8.34 & - \\
      OccWorld-D & 3D Occ &11.55 & 8.66 & 6.98 & 8.66 & 18.90 & 16.26 & 14.43 & 16.53  & - \\
      OccLLaMA-F & 3D Occ &10.34 &8.66 &6.98 &8.66 &\textbf{25.81} &\textbf{23.19} &\underline{19.97} &\textbf{22.99} & -\\
      \hline
      PreWorld & 3D Occ &11.69 & 8.72 & 6.77 & 9.06 & 23.01 & 20.79 & 18.84 & 20.88 & 1.0 \\
      +Pre-training & 2D \& 3D Occ &\underline{12.27} & \underline{9.24} & \underline{7.15} & \underline{9.55} & \underline{23.62} & 21.76 & 19.63 & 21.62 & 1.0 \\
      \hline
      \textbf{SparseWorld (Ours)} & 3D Occ  &\textbf{14.93} & \textbf{13.15} & \textbf{11.51} & \textbf{13.20} & 22.96 & \underline{22.10} & \textbf{21.05} & \underline{22.03} & \textbf{8.0} \\
      \hline
    \end{tabular*}
  }
  \caption{4D occupancy forecasting performance of Occ3D-nuScenes dataset. The best results are highlighted \textbf{bolded}, while the second-best results are \underline{underlined}. All comparative results are copied from the original papers for fairness.}
  \label{tab:result-forecasting}
\end{table*}

\begin{table*}[ht]
  \small
  \renewcommand{\arraystretch}{1.3}
  \makebox[\textwidth][c]{%
    \begin{tabular*}{\textwidth}{>{\centering\arraybackslash}m{3.0cm}|
    >{\centering\arraybackslash}m{4.8cm}|
    >{\centering\arraybackslash}m{0.8cm}
    >{\centering\arraybackslash}m{0.8cm}
    >{\centering\arraybackslash}m{0.8cm}
    >{\centering\arraybackslash}m{0.8cm}|
    >{\centering\arraybackslash}m{0.8cm}
    >{\centering\arraybackslash}m{0.8cm}
    >{\centering\arraybackslash}m{0.8cm}
    >{\centering\arraybackslash}m{0.8cm}}
      \hline
      \multirow{2}{*}{Method} & \multirow{2}{*}{Aux. Sup.} & \multicolumn{4}{c|}{L2 (m) $\downarrow$} & \multicolumn{4}{c}{Collision Rate (\%) $\downarrow$}  \\
      & & 1s & 2s & 3s & Avg. & 1s & 2s & 3s & Avg. \\
      \hline
      UniAD & Map\& Box\& Motion\& Track\& Occ &0.48 &0.96 &1.65& 1.03 &\textbf{0.05}& \textbf{0.17} &0.71& \underline{0.31}\\
       OccNet & Map \& Box \& 3D Occ &1.29 &2.13 &2.99 &2.14 &0.21 &0.59 &1.37 &0.72  \\
       \hline
      OccWorld-D & 3D Occ &0.52 & 1.27 & 2.41 & 1.40 & 0.12 & 0.40 & 2.08 & 0.87  \\
      OccLLaMA-F $\dagger$ &3D Occ &0.38 &1.07 &2.15 & 1.20 &0.06 &0.39 &1.65 & 0.70\\
      PreWorld & 2D \& 3D Occ &0.41 & 1.16 & 2.32 & 1.30 & 0.50 & 0.88 & 2.42 & 1.27  \\
      PreWorld $^\dagger$ & 2D \& 3D Occ &\underline{0.22} & \underline{0.30} & \underline{0.40} & \underline{0.31} & 0.21 & 0.66 & \underline{0.71} & 0.53 \\
      \hline
      \textbf{SparseWorld (Ours)} & 3D Occ  &0.49 & 0.94 & 1.47 & 0.97 & 0.18 &0.78 & 1.88 & 0.95  \\
      \textbf{SparseWorld$^\dagger$ (Ours)} & 3D Occ  &\textbf{0.19} & \textbf{0.25} & \textbf{0.36} & \textbf{0.27} & \underline{0.11} &\underline{0.29} & \textbf{0.46} &  \textbf{0.29} \\
      \hline
    \end{tabular*}
  }
  \caption{Motion planning performance on the Occ3D-nuScenes dataset. 
  $\dagger$ indicates that ego state is applied during training and inference. 
  The best results are in \textbf{bold}, second-best are \underline{underlined}.}
  \label{tab:result-planning}
\end{table*}

However, the timestamps of initial queries are discrete and can not be directly learned. There are two straightforward solutions:
(1) Manually assigning timestamps to queries and applying explicit supervision for each frame. However, this approach fails to provide effective supervision signals, due to the inherent spatial overlap between adjacent frames. (2) Ignoring temporal distinctions and using all queries to forecast all future frames. Yet this leads to convergence conflicts during training and ultimately degrades performance.

To address this challenge, we craft a Timestamp-Aware Self-Scheduling Training strategy. Specifically, we first pretrain RAP without explicitly assigning query timestamps while temporarily removing the temporal component in Eq. \ref{eq: attention weight}. The loss during pretraining is defined as:
\begin{equation}
    \mathcal{L}_{pretrain} = \text{CD}(P_0',\hat{\mathcal{G}}) + \sum_{l=1}^{L}(\text{CD}(P_l,\hat{\mathcal{G}}) +\mathcal{L}_\text{focal}(C_l,C_g)),
\end{equation}
where $C_g$ denotes semantic labels, $P_l$ denotes point set output by the $l$-th decoder. $\mathcal{L}_\text{focal}$ denotes the Focal Loss.

We construct an statistical matrix $M\in \mathbb{R}^{N\times(f+1)}$ that records the count of output points from each of $N$ queries corresponding to each timestamp over the entire dataset.

The ground truth timestamps are generated alongside $\hat{\mathcal{G}}$. As described in Eq. \ref{eq: chamfer}, if a predicted point $p$ is matched to a ground-truth point $g$ at time step 
$t$, the statistical counter of the source query of $p$ corresponding to $t$ is incremented by 1. Note that during the re-voxelization process, a single voxel $g$ may correspond to multiple timestamps.

Based on $M$, we further design a max-proportion prioritized assigning algorithm to selectively assign query timestamps, with details in the Appendix.

After 6 epochs of pretraining, the 3D positions and timestamps of initial queries stabilize. We then perform end-to-end training, during which the statistical matrix $M$ and query timestamps are updated dynamically each epoch. We employ the Chamfer distance and L2 loss to supervise the forecasted occupancy and trajectories, respectively. 

The total loss during end-to-end training is:
\begin{equation}
    \begin{aligned}
    \mathcal{L}_\text{total} = &\mathcal{L}_{pretrain}+\sum_{t=1}^{f}(\lambda_1 \text{CD}(P^t,\hat{\mathcal{G}}^t)\\
    &+\lambda_2\mathcal{L}_\text{focal}(C^t,\hat{C}^t)+\lambda_3\mathcal{L}_{2}(w^t,\hat{w}^t))
    \end{aligned}
\end{equation}
Here, $\hat{\mathcal{G}}^t$ and $\hat{w}^t$ denote the occupancy and trajectory ground truths at frame $t$, respectively. 

Notably, during inference, the query timestamps remain fixed, eliminating the query assigning process and ensuring the efficiency of SparseWorld.

\begin{figure*}
    
    \includegraphics[width=1.0\linewidth]{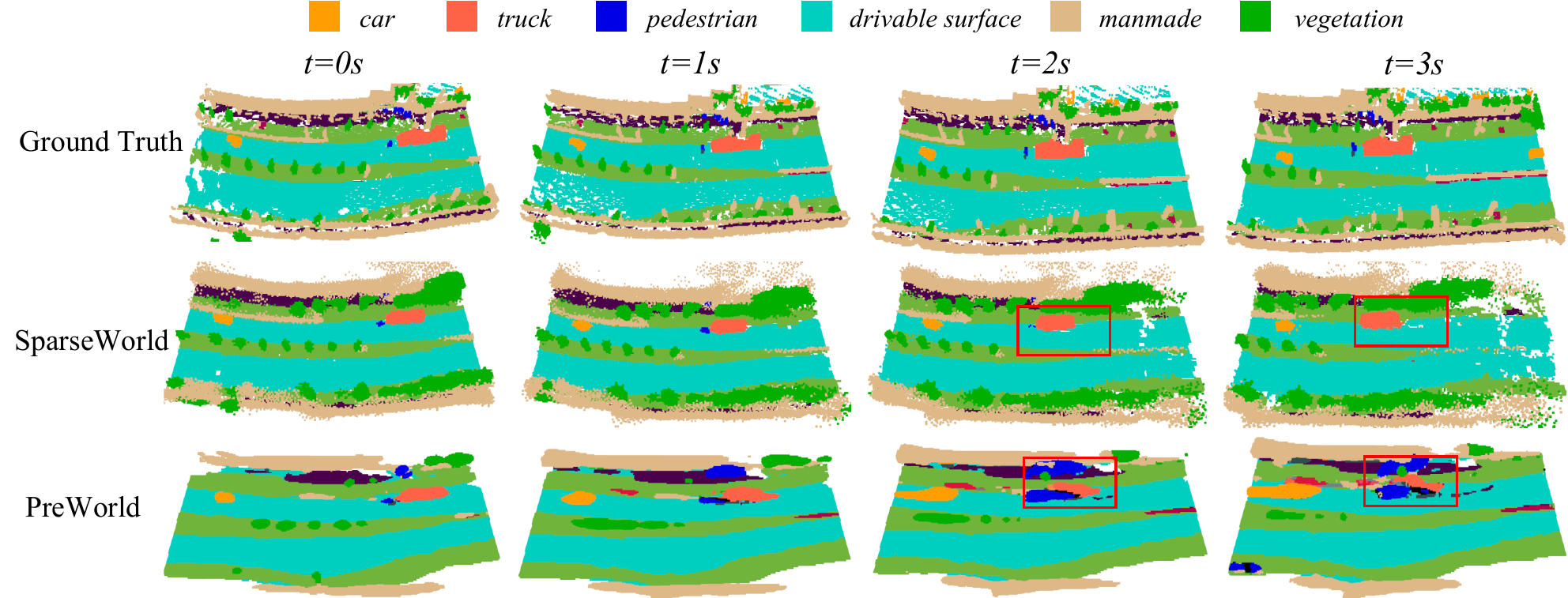}
    \caption{Visualization of the current and future 3-second ground truth and forecasts. As highlighted in red boxes, PreWorld \cite{preworld}, which relies on grid-based modeling and voxel-level classification, exhibits severe distortions in long horizon, whereas SparseWorld effectively mitigates such issues.}
    \label{fig:vis}
\end{figure*}
\section{Experiments}
\subsection{Experiment Settings}
\paragraph{Dataset and Metrics}
We employ the widely adopted Occ3d-nuScenes \cite{occ3d} benchmark, which is built upon the nuScenes dataset \cite{nuscenes}. It provides 700 training scenes and 150 validation scenes, each lasting 20 seconds, with labels provided every 0.5 seconds. The Occ3d-nuScenes dataset offers dense labels with a resolution of 200×200×16, comprising 17 semantic categories and one free category. Each occupancy grid cell has a size of 0.4m×0.4m×0.4m. Following established methodologies, we assess occupancy perception and forecasting using Intersection over Union (IoU) and mean IoU (mIoU) as metrics. IoU evaluates overleap considering only foreground and background, whereas mIoU computes the mean IoU across all 17 classes. L2 error and collision rate are applied as indicators for ego-vehicle trajectory planning.

\paragraph{Implementation Details}
In the RAP module, We employ 6 decoding layers, where the number of output points for each query across successive layers is (1, 4, 16, 24, 32, 48). The number of queries for the 7 time steps (the current and 6 future frames) is divided as (720, 60, 60, 60, 60, 40, 40), resulting in a total of 1040 initial queries. In our implementation, we utilize ResNet-50 \cite{resnet} with 256$\times$704 images to extract multi-scale features. 

We train our model using Temporal-Aware Self-Scheduling strategy and the AdamW optimizer, with a standard learning rate set to 2e-4. A warm-up strategy and cosine annealing mechanism are applied. The model undergoes 6 epochs of pre-training, followed by 48 epochs of end-to-end training. The entire training is carried out on 4 A100 GPUs, with a total batch size of 8. The inference speed is measured on a 4090 GPU. For further implementation details, please refer to the appendix. Notably, the visible masks are \textbf{not} utilized during both training and inference.

\subsection{Main Results}
Following established methodologies, we take the current and past 2 seconds of video frames as input to forecast the occupancy and ego-vehicle trajectory for the next 3 seconds. 

\paragraph{4D Occupancy Forecasting}

Table \ref{tab:result-forecasting} compares the performance of SparseWorld with other excellent occupancy world models in terms of 4D forecasting. SparseWorld demonstrates exceptional performance, achieving a 20\%-40\% improvement in mIoU over PreWorld \cite{preworld}, along with a 7$\times$ increase in inference speed. Notably, SparseWorld exhibits the smallest score degradation during autoregressive forecasting among all models, highlighting the substantial advantages of dynamic and continuous queries in forecasting tasks. However, SparseWorld does not show a significant advantage on the IoU metric. We guess this is because IoU is dominated by background voxels (accounting for roughly 95\%), most of which are not visible, whereas SparseWorld excels at recognizing foreground objects.

Our visualization of the inference results in Figure \ref{fig:vis} further supports this conclusion. We observed that grid feature-based methods encounter feature distortion during frame-by-frame forecasting, leading to significant accumulative errors, especially in foreground categories that require particular attention. Our SparseWorld effectively avoids these issues. More visual examples can be found in the appendix.

\paragraph{Motion Planning}
Table \ref{tab:result-planning} presents the motion planning results of SparseWorld. Clearly, our model also demonstrates excellent trajectory planning capabilities, particularly in terms of collision rates, where SparseWorld consistently achieves only half the collision rate of PreWorld. To ensure a fair comparison, we do not use the vehicle's state when generating the ego token, yet the model's performance remains impressive. We attribute this to the inherent nature of dynamic and continuous regression, which contributes to superior perception and 4D forecasting abilities, laying a solid foundation for safe and reasonable path planning.

\subsection{Ablation Studies}

\begin{table}[]
\centering
\begin{tabular}{l|c|c}
\hline
  Module  & Avg. mIoU& Avg. IoU \\
\hline
 SparseWorld & \textbf{11.82} & \textbf{21.17} \\
 w/o Adaptive Scaling &  11.45 (-0.37) & 20.88 (-0.29)\\
 w/o Temporal mask & 11.52 (-0.3) & 20.89 (-0.28) \\
 w/o 4D PE & 11.58 (-0.24) & 20.99 (-0.18) \\
 w/o State Condition & 11.31 (-0.51) & 20.62 (-0.55)\\

\hline
\end{tabular}
\caption{Ablation studies of core model components.}
\label{tab:ablation}
\end{table}

\begin{figure*}
    \centering
    \includegraphics[width=1.0\linewidth]{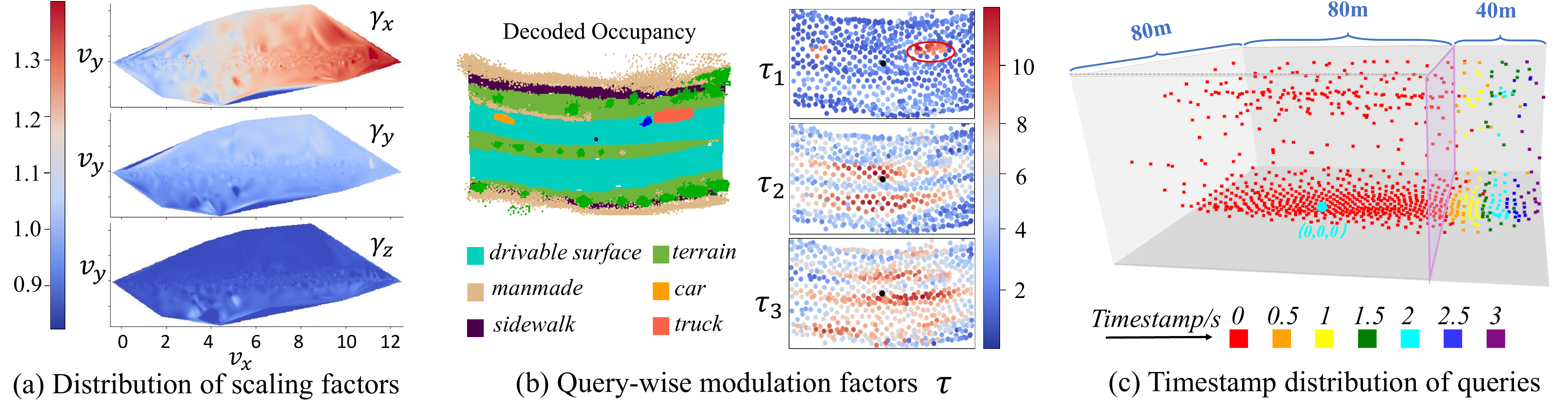}
    \caption{Visualizations of intermediate reasoning representations.}
    \label{fig:intermediate}
\end{figure*}

In this section, we conduct detailed ablation experiments to investigate the impact of various designs on model and further validate our arguments. To expedite the verification process, we assume the use of SparseWorld, pre-training for 6 epochs and fully training the model 12 epochs, respectively.

\paragraph{Model Components}
Table \ref{tab:ablation} presents the ablation study evaluating the contribution of core modules. We report the average IoU and average mIoU within the future 3 seconds.

As shown in Row 3 of Table \ref{tab:ablation}, the Adaptive Scaling improves mIoU by 0.37, demonstrating the significance of adaptive perception range for motion-aware scene forecasting. In Figure \ref{fig:intermediate}(a), we further visualize the heatmaps of learned scaling factors $\bm{\gamma}$ under different ego-motion states across the nuScenes validation set. It is observed that the longitudinal velocity $v_x$ exhibits a larger value range than $v_y$ and remains positive (indicating no reverse motion). The scaling factor $\gamma_x$ is more sensitive to $v_x$, while $\gamma_y$ and $\gamma_z$ exhibit no significant differences across ego states. This confirms that larger perception ranges are required in the longitudinal direction, which aligns with the motion prior in autonomous driving.

As shown in row 4 of Table \ref{tab:ablation}, without the temporal mask of Temporal-Spatial MHSA, SparseWorld suffered a performance loss of 0.3. This is because, although adaptive perception is necessary, the range of ground truth for each frame is fixed ($\pm 40$m in Occ3D-nuScenes). Future queries can ``mislead" the current query, thus disrupting the completion of the current scene. Conversely, subsequent frames can attend to the previous frame's query, allowing later frames to supplement earlier ones .

As shown in row 5 of Table \ref{tab:ablation}, the 4D position encoding in SCF resulted in a performance improvement of 0.24. We believe that the introduction of spatial position allows the model to learn to correct potential perceptual errors, while the temporal information enables the model to consciously adjust the magnitude of refinement.

As shown in row 6 of Table \ref{tab:ablation}, removing the Ego State Condition leads to a substantial drop of 0.51 mIoU, highlighting the critical role of ego motion in forecasting.



\begin{table}[]
\centering
\begin{tabular}{c|c|c|c|c}
\hline
  Ego  &  Cross-Attn  & L2 (m) $\downarrow$ & Col (\%) $\downarrow$ & mIoU $\uparrow$ \\
\hline  
 $\checkmark$ & Spatial & 0.29 & 0.37& 11.82\\
  & Spatial & 1.01 & 0.65& 10.64\\
 $\checkmark$ & Common & 0.31 & 0.44 &  11.71\\
   & Common & 1.25 & 0.77 & 10.28  \\
\hline
\end{tabular}
\caption{Ablation studies on ego-state and spatial cross-attention, we report the average mIoU over future 3 seconds.}
\label{tab:planning}
\end{table}

\paragraph{Ego status and Spatial Cross-attention}
In Table \ref{tab:planning}, we investigate the impact of ego-state spatial cross-attention on planning. Clearly, better planning consistently promotes more accurate occupancy forecast. This is intuitive: as larger trajectory errors hinder the model from capturing scene shifts caused by ego-motion. Moreover, compared to common cross-attention, introducing spatial modulation significantly improves planning performance, particularly in the absence of ego-state input.

To better understand this, we visualize heatmaps of several heads of $\tau$ values within a given scene in Figure \ref{fig:intermediate}(b), where higher values indicate stronger attention from the ego agent. We observe that $\tau_1$ assigns higher weights to foreground queries, while $\tau_2$ focuses more on queries corresponding to drivable surfaces. This suggests that adaptive modulation enables the model to integrate both foreground and background cues when making planning decisions.

\paragraph{Temporal-Aware Self-Scheduling Training}
Training SparseWorld, a multi-stage and multi-output model, is non-trivial.  In Table \ref{tab:training}, we compare different training strategies to investigate the effect of Temporal-Aware Self-Scheduling. Without temporal differentiation, where all queries are supervised by multiple frames of GT, the training suffers from convergence instability, leading to a catastrophic drop of 0.87 mIoU. When timestamps are manually assigned, each perception decoder must learn positions under $f$-time supervisions, requiring $L\times f$ Chamfer distance computations, which is computationally inefficient and yields suboptimal results. In contrast, our Temporal-Aware Self-Scheduling only requires $L+f$ Chamfer distance computations. And through pretraining, the model learns the temporal distribution of queries on its own, enabling more efficient and smoother convergence. 

In Figure 4(c), we visualize the learned queries with timestamp distributions, and distinguish them by color. It obviously that the queries exhibit a clear hierarchical structure along the longitudinal direction, which benefits long-range perception and facilitates cross-temporal forecasting.

\begin{table}[]
\centering
\begin{tabular}{c|c|c|c}
\hline
  Training Strategy  &  Avg. mIoU & Avg. IoU & Time \\
\hline
 Tem-Aware Self-Sche &   11.82 & 21.17 &12h \\
 No Tem Different &   10.95 & 20.08 & 11h \\
 Manually Specified  & 11.37 & 20.31 & 22h \\

\hline
\end{tabular}
\caption{Ablation results of different training strategies.}
\label{tab:training}
\end{table}

\section{Conclusion}
In this paper, we proposed SparseWorld, a fully sparse, flexible, adaptive, and efficient 4D occupancy world model. Guided by ego state, SparseWorld achieves adaptive-range perception and ego-conditioned forecasting through bidirectional query interaction, aligning with the continuous dynamics of 4D scenes. A dedicated Temporal-Aware Self-Scheduling strategy further ensures stable training.

Extensive experiments demonstrate SparseWorld's superior forecasting and planning capabilities, while ablation studies and visualizations confirm its interpretability.
\subsection{Limitations and Future work}
Despite its strong performance, SparseWorld’s geometric reasoning (IoU) and generalization to unseen scenarios remain areas for improvement. Notably, it requires only LiDAR and 2D labels without dense occupancy annotations. Future work will explore weakly supervised training and integrate large language models to enhance reasoning in complex scenes. Overall, SparseWorld demonstrates that a small number of sparse queries can effectively represent 4D scenes, offering a new perspective for autonomous driving research.

\section{Acknowledgements}
This work is funded by Xiongan AI Institute, Lenovo Research and Wuxi Research Institute of Applied Technologies, Tsinghua University under Grant 20242001120. We sincerely appreciate their supports and contributions.

\bibliography{aaai2026}

@inproceedings{occworld,
  title={Occworld: Learning a 3d occupancy world model for autonomous driving},
  author={Zheng, Wenzhao and Chen, Weiliang and Huang, Yuanhui and Zhang, Borui and Duan, Yueqi and Lu, Jiwen},
  booktitle={European conference on computer vision},
  pages={55--72},
  year={2024},
  organization={Springer}
}

@article{occllama,
  title={Occllama: An occupancy-language-action generative world model for autonomous driving},
  author={Wei, Julong and Yuan, Shanshuai and Li, Pengfei and Hu, Qingda and Gan, Zhongxue and Ding, Wenchao},
  journal={arXiv preprint arXiv:2409.03272},
  year={2024}
}

@article{preworld,
  title={Semi-Supervised Vision-Centric 3D Occupancy World Model for Autonomous Driving},
  author={Li, Xiang and Li, Pengfei and Zheng, Yupeng and Sun, Wei and Wang, Yan and Chen, Yilun},
  journal={arXiv preprint arXiv:2502.07309},
  year={2025}
}

@inproceedings{driveoccworld,
  title={Driving in the occupancy world: Vision-centric 4d occupancy forecasting and planning via world models for autonomous driving},
  author={Yang, Yu and Mei, Jianbiao and Ma, Yukai and Du, Siliang and Chen, Wenqing and Qian, Yijie and Feng, Yuxiang and Liu, Yong},
  booktitle={Proceedings of the AAAI Conference on Artificial Intelligence},
  volume={39},
  number={9},
  pages={9327--9335},
  year={2025}
}

@article{opus,
  title={Opus: occupancy prediction using a sparse set},
  author={Wang, Jiabao and Liu, Zhaojiang and Meng, Qiang and Yan, Liujiang and Wang, Ke and Yang, Jie and Liu, Wei and Hou, Qibin and Cheng, Ming-Ming},
  journal={arXiv preprint arXiv:2409.09350},
  year={2024}
}

@InProceedings{sparsebev,
    author    = {Liu, Haisong and Teng, Yao and Lu, Tao and Wang, Haiguang and Wang, Limin},
    title     = {SparseBEV: High-Performance Sparse 3D Object Detection from Multi-Camera Videos},
    booktitle = {Proceedings of the IEEE/CVF International Conference on Computer Vision (ICCV)},
    month     = {October},
    year      = {2023},
    pages     = {18580-18590}
}

@article{occ3d,
  title={Occ3d: A large-scale 3d occupancy prediction benchmark for autonomous driving},
  author={Tian, Xiaoyu and Jiang, Tao and Yun, Longfei and Mao, Yucheng and Yang, Huitong and Wang, Yue and Wang, Yilun and Zhao, Hang},
  journal={Advances in Neural Information Processing Systems},
  volume={36},
  pages={64318--64330},
  year={2023}
}

@inproceedings{nuscenes,
  title={nuscenes: A multimodal dataset for autonomous driving},
  author={Caesar, Holger and Bankiti, Varun and Lang, Alex H and Vora, Sourabh and Liong, Venice Erin and Xu, Qiang and Krishnan, Anush and Pan, Yu and Baldan, Giancarlo and Beijbom, Oscar},
  booktitle={Proceedings of the IEEE/CVF conference on computer vision and pattern recognition},
  pages={11621--11631},
  year={2020}
}

@inproceedings{resnet,
  title={Deep residual learning for image recognition},
  author={He, Kaiming and Zhang, Xiangyu and Ren, Shaoqing and Sun, Jian},
  booktitle={Proceedings of the IEEE conference on computer vision and pattern recognition},
  pages={770--778},
  year={2016}
}

@inproceedings{sparseocc,
  title={Fully sparse 3d occupancy prediction},
  author={Liu, Haisong and Chen, Yang and Wang, Haiguang and Yang, Zetong and Li, Tianyu and Zeng, Jia and Chen, Li and Li, Hongyang and Wang, Limin},
  booktitle={European Conference on Computer Vision},
  pages={54--71},
  year={2024},
  organization={Springer}
}

@inproceedings{osp,
  title={Occupancy as set of points},
  author={Shi, Yiang and Cheng, Tianheng and Zhang, Qian and Liu, Wenyu and Wang, Xinggang},
  booktitle={European Conference on Computer Vision},
  pages={72--87},
  year={2024},
  organization={Springer}
}

@article{flashocc,
  title={Flashocc: Fast and memory-efficient occupancy prediction via channel-to-height plugin},
  author={Yu, Zichen and Shu, Changyong and Deng, Jiajun and Lu, Kangjie and Liu, Zongdai and Yu, Jiangyong and Yang, Dawei and Li, Hui and Chen, Yan},
  journal={arXiv preprint arXiv:2311.12058},
  year={2023}
}

@inproceedings{cotr,
  title={Cotr: Compact occupancy transformer for vision-based 3d occupancy prediction},
  author={Ma, Qihang and Tan, Xin and Qu, Yanyun and Ma, Lizhuang and Zhang, Zhizhong and Xie, Yuan},
  booktitle={Proceedings of the IEEE/CVF Conference on Computer Vision and Pattern Recognition},
  pages={19936--19945},
  year={2024}
}

@inproceedings{panoocc,
  title={Panoocc: Unified occupancy representation for camera-based 3d panoptic segmentation},
  author={Wang, Yuqi and Chen, Yuntao and Liao, Xingyu and Fan, Lue and Zhang, Zhaoxiang},
  booktitle={Proceedings of the IEEE/CVF conference on computer vision and pattern recognition},
  pages={17158--17168},
  year={2024}
}

@article{lightocc,
  title={Lightweight Spatial Embedding for Vision-based 3D Occupancy Prediction},
  author={Zhang, Jinqing and Zhang, Yanan and Liu, Qingjie and Wang, Yunhong},
  journal={arXiv preprint arXiv:2412.05976},
  year={2024}
}

@article{heightocc,
  title={Deep height decoupling for precise vision-based 3d occupancy prediction},
  author={Wu, Yuan and Yan, Zhiqiang and Wang, Zhengxue and Li, Xiang and Hui, Le and Yang, Jian},
  journal={arXiv preprint arXiv:2409.07972},
  year={2024}
}

@inproceedings{protoocc,
  title={Protoocc: Accurate, efficient 3d occupancy prediction using dual branch encoder-prototype query decoder},
  author={Kim, Jungho and Kang, Changwon and Lee, Dongyoung and Choi, Sehwan and Choi, Jun Won},
  booktitle={Proceedings of the AAAI Conference on Artificial Intelligence},
  volume={39},
  number={4},
  pages={4284--4292},
  year={2025}
}

@article{bevdet4d,
  title={Bevdet4d: Exploit temporal cues in multi-camera 3d object detection},
  author={Huang, Junjie and Huang, Guan},
  journal={arXiv preprint arXiv:2203.17054},
  year={2022}
}

@article{sun2024gsrender,
  title={Gsrender: Deduplicated occupancy prediction via weakly supervised 3d gaussian splatting},
  author={Sun, Qianpu and Shu, Changyong and Zhou, Sifan and Yu, Zichen and Chen, Yan and Yang, Dawei and Chun, Yuan},
  journal={arXiv preprint arXiv:2412.14579},
  year={2024}
}

@inproceedings{selfocc,
  title={Selfocc: Self-supervised vision-based 3d occupancy prediction},
  author={Huang, Yuanhui and Zheng, Wenzhao and Zhang, Borui and Zhou, Jie and Lu, Jiwen},
  booktitle={Proceedings of the IEEE/CVF Conference on Computer Vision and Pattern Recognition},
  pages={19946--19956},
  year={2024}
}

@inproceedings{renderocc,
  title={Renderocc: Vision-centric 3d occupancy prediction with 2d rendering supervision},
  author={Pan, Mingjie and Liu, Jiaming and Zhang, Renrui and Huang, Peixiang and Li, Xiaoqi and Xie, Hongwei and Wang, Bing and Liu, Li and Zhang, Shanghang},
  booktitle={2024 IEEE International Conference on Robotics and Automation (ICRA)},
  pages={12404--12411},
  year={2024},
  organization={IEEE}
}

@article{gaussianflowocc,
  title={GaussianFlowOcc: Sparse and Weakly Supervised Occupancy Estimation using Gaussian Splatting and Temporal Flow},
  author={Boeder, Simon and Gigengack, Fabian and Risse, Benjamin},
  journal={arXiv preprint arXiv:2502.17288},
  year={2025}
}

@article{worldmodels,
  title={World models},
  author={Ha, David and Schmidhuber, J{\"u}rgen},
  journal={arXiv preprint arXiv:1803.10122},
  year={2018}
}

@article{dome,
  title={Dome: Taming diffusion model into high-fidelity controllable occupancy world model},
  author={Gu, Songen and Yin, Wei and Jin, Bu and Guo, Xiaoyang and Wang, Junming and Li, Haodong and Zhang, Qian and Long, Xiaoxiao},
  journal={arXiv preprint arXiv:2410.10429},
  year={2024}
}

@inproceedings{voxformer,
  title     = {VoxFormer: Sparse Voxel Transformer for Camera-based 3D Semantic Scene Completion},
  author    = {Li, Yiyi and Yu, Zhiqiang and Choy, Christopher and et al.},
  booktitle = {Proceedings of the IEEE/CVF Conference on Computer Vision and Pattern Recognition (CVPR)},
  pages     = {9087--9098},
  year      = {2023}
}

@inproceedings{cvtocc,
  title={Cvt-occ: Cost volume temporal fusion for 3d occupancy prediction},
  author={Ye, Zhangchen and Jiang, Tao and Xu, Chenfeng and Li, Yiming and Zhao, Hang},
  booktitle={European Conference on Computer Vision},
  pages={381--397},
  year={2024},
  organization={Springer}
}

@article{occllm,
  title={Occ-llm: Enhancing autonomous driving with occupancy-based large language models},
  author={Xu, Tianshuo and Lu, Hao and Yan, Xu and Cai, Yingjie and Liu, Bingbing and Chen, Yingcong},
  journal={arXiv preprint arXiv:2502.06419},
  year={2025}
}

@article{magicdrive,
  title={Magicdrive: Street view generation with diverse 3d geometry control},
  author={Gao, Ruiyuan and Chen, Kai and Xie, Enze and Hong, Lanqing and Li, Zhenguo and Yeung, Dit-Yan and Xu, Qiang},
  journal={arXiv preprint arXiv:2310.02601},
  year={2023}
}

@article{vista,
  title={Vista: A generalizable driving world model with high fidelity and versatile controllability},
  author={Gao, Shenyuan and Yang, Jiazhi and Chen, Li and Chitta, Kashyap and Qiu, Yihang and Geiger, Andreas and Zhang, Jun and Li, Hongyang},
  journal={Advances in Neural Information Processing Systems},
  volume={37},
  pages={91560--91596},
  year={2024}
}

@inproceedings{vidar,
  title={Visual point cloud forecasting enables scalable autonomous driving},
  author={Yang, Zetong and Chen, Li and Sun, Yanan and Li, Hongyang},
  booktitle={Proceedings of the IEEE/CVF Conference on Computer Vision and Pattern Recognition},
  pages={14673--14684},
  year={2024}
}

@inproceedings{stp3,
  title={St-p3: End-to-end vision-based autonomous driving via spatial-temporal feature learning},
  author={Hu, Shengchao and Chen, Li and Wu, Penghao and Li, Hongyang and Yan, Junchi and Tao, Dacheng},
  booktitle={European Conference on Computer Vision},
  pages={533--549},
  year={2022},
  organization={Springer}
}

@inproceedings{uniad,
  title={Planning-oriented autonomous driving},
  author={Hu, Yihan and Yang, Jiazhi and Chen, Li and Li, Keyu and Sima, Chonghao and Zhu, Xizhou and Chai, Siqi and Du, Senyao and Lin, Tianwei and Wang, Wenhai and others},
  booktitle={Proceedings of the IEEE/CVF conference on computer vision and pattern recognition},
  pages={17853--17862},
  year={2023}
}

@inproceedings{vad,
  title={Vad: Vectorized scene representation for efficient autonomous driving},
  author={Jiang, Bo and Chen, Shaoyu and Xu, Qing and Liao, Bencheng and Chen, Jiajie and Zhou, Helong and Zhang, Qian and Liu, Wenyu and Huang, Chang and Wang, Xinggang},
  booktitle={Proceedings of the IEEE/CVF International Conference on Computer Vision},
  pages={8340--8350},
  year={2023}
}

@article{sparsead,
  title={Sparsead: Sparse query-centric paradigm for efficient end-to-end autonomous driving},
  author={Zhang, Diankun and Wang, Guoan and Zhu, Runwen and Zhao, Jianbo and Chen, Xiwu and Zhang, Siyu and Gong, Jiahao and Zhou, Qibin and Zhang, Wenyuan and Wang, Ningzi and others},
  journal={arXiv preprint arXiv:2404.06892},
  year={2024}
}

@article{sparsedrive,
  title={Sparsedrive: End-to-end autonomous driving via sparse scene representation},
  author={Sun, Wenchao and Lin, Xuewu and Shi, Yining and Zhang, Chuang and Wu, Haoran and Zheng, Sifa},
  journal={arXiv preprint arXiv:2405.19620},
  year={2024}
}

@article{drivetransformer,
  title={Drivetransformer: Unified transformer for scalable end-to-end autonomous driving},
  author={Jia, Xiaosong and You, Junqi and Zhang, Zhiyuan and Yan, Junchi},
  journal={arXiv preprint arXiv:2503.07656},
  year={2025}
}

@article{fbocc,
  title={Fb-occ: 3d occupancy prediction based on forward-backward view transformation},
  author={Li, Zhiqi and Yu, Zhiding and Austin, David and Fang, Mingsheng and Lan, Shiyi and Kautz, Jan and Alvarez, Jose M},
  journal={arXiv preprint arXiv:2307.01492},
  year={2023}
}

@article{vadv2,
  title={Vadv2: End-to-end vectorized autonomous driving via probabilistic planning},
  author={Chen, Shaoyu and Jiang, Bo and Gao, Hao and Liao, Bencheng and Xu, Qing and Zhang, Qian and Huang, Chang and Liu, Wenyu and Wang, Xinggang},
  journal={arXiv preprint arXiv:2402.13243},
  year={2024}
}

@inproceedings{diffusiondrive,
  title={Diffusiondrive: Truncated diffusion model for end-to-end autonomous driving},
  author={Liao, Bencheng and Chen, Shaoyu and Yin, Haoran and Jiang, Bo and Wang, Cheng and Yan, Sixu and Zhang, Xinbang and Li, Xiangyu and Zhang, Ying and Zhang, Qian and others},
  booktitle={Proceedings of the Computer Vision and Pattern Recognition Conference},
  pages={12037--12047},
  year={2025}
}

@article{diffusion-planner,
  title={Diffusion-based planning for autonomous driving with flexible guidance},
  author={Zheng, Yinan and Liang, Ruiming and Zheng, Kexin and Zheng, Jinliang and Mao, Liyuan and Li, Jianxiong and Gu, Weihao and Ai, Rui and Li, Shengbo Eben and Zhan, Xianyuan and others},
  journal={arXiv preprint arXiv:2501.15564},
  year={2025}
}

\end{document}